\documentclass[letterpaper, 10 pt, conference]{ieeeconf}

\IEEEoverridecommandlockouts                              

\usepackage[noformatting,nogeometry]{stdcommands}
\usepackage{graphicx}
\usepackage{dsfont} 
\usepackage{amsmath}
\usepackage{amssymb}
\usepackage{url}
\usepackage[utf8]{inputenc} 
\usepackage[format=hang,singlelinecheck=false,caption=false,font=footnotesize]{subfig}
\usepackage{changes}
\usepackage[ruled,linesnumbered]{algorithm2e}
\usepackage{scalerel}
\usepackage[noend]{algpseudocode}
\usepackage[english]{babel}
\usepackage{mathrsfs}
\usepackage{comment}
\usepackage{hhline}
\graphicspath{images}

\title{\LARGE \bf DL-IAPS and PJSO: A Path/Speed Decoupled Trajectory Optimization and its Application in Autonomous Driving}

\begin{document}

\author{%
    \parbox{\linewidth}{\centering
      Jinyun Zhou$^{1}$,
      Runxin He$^{1}$,
      Yu Wang,
      Shu Jiang,
      Zhenguang Zhu,
      Jiangtao Hu,
      Jinghao Miao
      and Qi Luo$^{2}$
  }%
  \thanks{$^1$ Authors contributed equally to this paper}
  \thanks{$^2$ Corresponding author}
  \thanks{All Authors are with Baidu USA LLC,
        250 E Caribbean Drive, Sunnyvale, CA 94089
        {\tt\small jinyunzhou@baidu.com, runxinhe@baidu.com, luoqi06@baidu.com}}%
}

\markboth{}{}

\maketitle
\thispagestyle{empty}
\pagestyle{empty}

\begin{abstract}
This paper presents a free space trajectory optimization algorithm of autonomous driving vehicle, which decouples the collision-free trajectory planning problem into a Dual-Loop Iterative Anchoring Path Smoothing (DL-IAPS) and a Piece-wise Jerk Speed Optimization (PJSO). The work leads to remarkable driving performance improvements including more precise collision avoidance, higher control feasibility and better driving comfort, as those are often hard to realize in other existing path/speed decoupled trajectory optimization methods.
Our algorithm's efficiency, robustness and adaptiveness to complex driving scenarios have been validated by both simulations and real on-road tests.
\end{abstract}

\IEEEpeerreviewmaketitle

\section{Introduction}
%
%
%
%

In recent years, autonomous driving technology is making a huge progress on handling numerous lane cruising scenarios including lane following, lane changing, stopping at traffic light, etc.~\cite{paden2016survey}\cite{schwarting2018planning}. However, many of the trajectory planning algorithms restrict the vehicle to follow the lane or disallow backward driving~\cite{Werling}. Such restrictions degrade the vehicle's capability to handle parallel and perpendicular parking or some scenarios when the car needs backward driving or going through a semi-structured area. So free space trajectory planning allowing both forward and backward gear is essential for expanding the vehicles' geo-fenced operation areas and enabling curb-to-curb operation. Free space trajectory planning algorithm for autonomous driving is required to consider non-holonomic vehicle dynamic constraints, exact obstacle collision avoidance and
real time computation, which make it a challenging and hot topic. Historically, two major approaches of general free space trajectory planning have been researched and applied to real test scenarios:

One category is path/speed coupled method which jointly solves the path and speed optimization based on nonlinear kinematic vehicle model. An example is Nonlinear Model Predictive Control (NMPC) framework~\cite{rosmann2017kinodynamic}\cite{zhang2018autonomous}. Some NMPC frameworks use Mixed Interger Programming (MIP) for obstacle avoidance constraint~\cite{MIP_application1} while others like Hierarchical Optimization-Based Collision Avoidance (H-OBCA) by X.Zhang use strong duality of convex optimization~\cite{zhang2018autonomous}. These approaches can elegantly incorporate both vehicle dynamics and obstacle avoidance into one single optimization problem but usually have high computation complexity and lower robustness~\cite{mayne2000constrained}\cite{mixedintegerreview}.

The other category is path/speed decoupled method that first smooths the path and then optimizes the speed profile along the path~\cite{kant1986decomposition}\cite{fraichard1993decomposition}.  This approach has better computational efficiency but usually lacks control feasibility and cannot guarantee path or speed smoothness in extreme cases~\cite{dolgov2008practical}\cite{aoude2010sampling}.

In this paper, we propose a novel path/speed decoupled method. On a basis of a global jerky and coarse reference path by searching based or sampling based path planning method (Hybrid A*~\cite{dolgov2010path}) in our case), we decouple the trajectory optimization problem into two hierarchical steps including, iterative curvature constrained path smoothing (i.e., DL-IAPS) and comfortable minimum-time piece-wise jerk speed optimization (i.e., PJSO). This addresses the above mentioned issues with following advantages:

\begin{enumerate}
	\item \textbf{Precise Collision Avoidance in Real-Time}:
    Existing works of obstacle and ego vehicle modeling~\cite{schulman2014motion}\cite{zhu2015convex}, either made a linear approximation to the collision avoidance constraint or approximated the ego vehicle's shape as a circle. The estimation error is difficult to evaluate; furthermore, some obstacle convexification work like approximating obstacles as a collection of circles in~\cite{Alrifaee} is difficult to gauge in a complex environment. In our DL-IAPS, we perform iterative collision checks with precise obstacle shapes and polygon-like vehicle shapes with an average time of 0.07s (with no obstacle) and about $0.18-0.21$s (with complex obstacles/boundaries) for complete trajectory (trajectory full-length is $9-14$s) as shown in~\ref{subsec:simple_simulation}. 
    
	\item \textbf{Control Feasibility}: 
	In order to speed up the trajectory generation, prior works either neglected to incorporate the maximum curvature/acceleration constraints introduced by non-holonomic vehicle dynamics~\cite{chen_liu_tomizuka_2018}\cite{cfs}, or as in the work named Convex Elastic Band Smoothing (CES) of Z. Zhu, only approximated maximum curvature constraint with certain assumptions~\cite{zhu2015convex}. These approaches may fail the constraint satisfaction in extreme cases, and hence degrade the control performance. Our approach overcomes this issue by strictly enforcing non-holonomic constraints with modified Sequential Convex Optimization (SCP) path planning, with results comparison shown in \ref{subsec:simple_simulation}.
	
	\item \textbf{Driving comfort and Minimum Traversal Time}: 
	T. Lipp and S. Boy proposed a minimum time profile generation with only time minimization as optimization objective~\cite{lipp_boyd_2014}. W. Lim et. al considered both minimum time and driving comfort in objective function but failed to include the hard constraints in speed profile optimization~\cite{WLim_2018}. We formulate the speed profile optimization in a form where both minimum traversal time and driving comfort are included in optimization objective and constraints, which provides better driving experience in autonomous Robotaxi application.
\end{enumerate}

The planner  is  integrated in  the  the  Apollo  Autonomous  Driving  Platform.\footnote{Source code available at ~\url{https://github.com/ApolloAuto/apollo}}.
We validated our work through 80 numeric simulation cases with different initial conditions, 208 simulation scenarios extracted from real world, and 400 hours on-road tests including US field tests and China T4-level tests~\cite{t4test} to confirm the efficiency and robustness of our proposed method in different free space driving scenarios.


This paper is organized as followed: the problem statement and the detailed description of the planning method are presented in Section~\ref{sec:problem_statement} and~\ref{sec:iterative_anchoring} respectively; the results of both simulations and on-road tests are presented in Section~\ref{sec:experiments}.

\section{Problem Statement}
\label{sec:problem_statement}
\begin{figure}[ht]
 \centering
\includegraphics[trim=0 0 0 0, clip, width=1.0\linewidth]{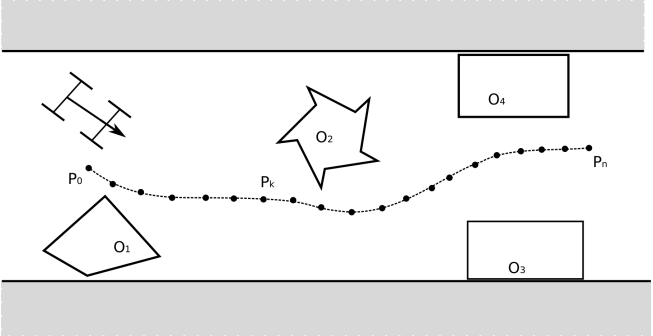}
 \caption{Illustration to problem statement}
 \label{figure:problem}
\end{figure}

\begin{figure}[ht]
 \centering
\includegraphics[trim=0 0 0 0, clip, width=1.0\linewidth]{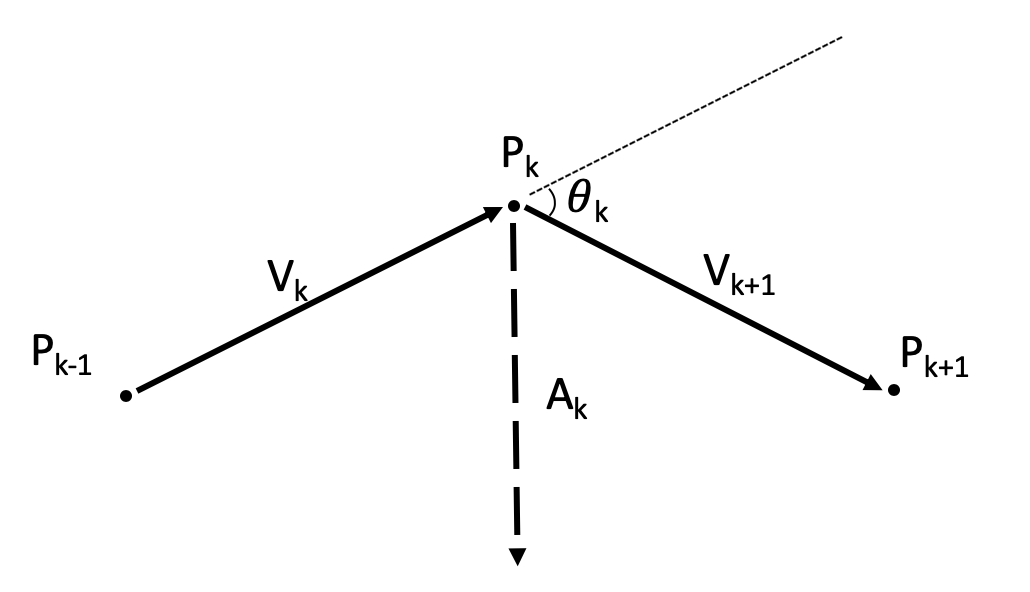}
 \caption{Illustration to path points notations $P_k$, $V_k$ and $A_k$}
 \label{figure:point_relation}
\end{figure}

As shown in Fig.~\ref{figure:problem}, $\mathcal{W} \subset \R^2$ denotes the work space for a vehicle, and $\mathcal{O} = \{O_i\}_{i=1}^m$ denotes the collection of obstacles.
At time step $k$, the ego vehicle's state can be described with location $P_k = [x_k, y_k]$, heading angle $\phi_k$ and unit heading vector $\hat{u}_{\phi_k}$.
Also, as shown in Fig.~\ref{figure:point_relation}, given two consecutive step $k-1$ and $k$, ego vehicle position change vector can be defined as $V_k = P_k - P_{k-1}$, and its change rate vector defined as $A_k = V_{k + 1} - V_{k}$, the included angle between two consecutive position change vector $V_{k+1} \text{ and } V_{k}$ is defined as  $\theta_{k}$. For point $P_k$, the longitudinal traverse distance, speed and acceleration are defined as $[s_k, \dot{s}_k, \ddot{s}_k] \in \R^3$.

With the proposed algorithm, a complete autonomous driving planning module architecture is designed and implemented, as shown in Fig.~\ref{figure:open_space_architect}. A trajectory planner, named Open Space Planner, contains three consecutive modules:

\begin{figure}[ht]
\centering
\includegraphics[width=1.0\linewidth]{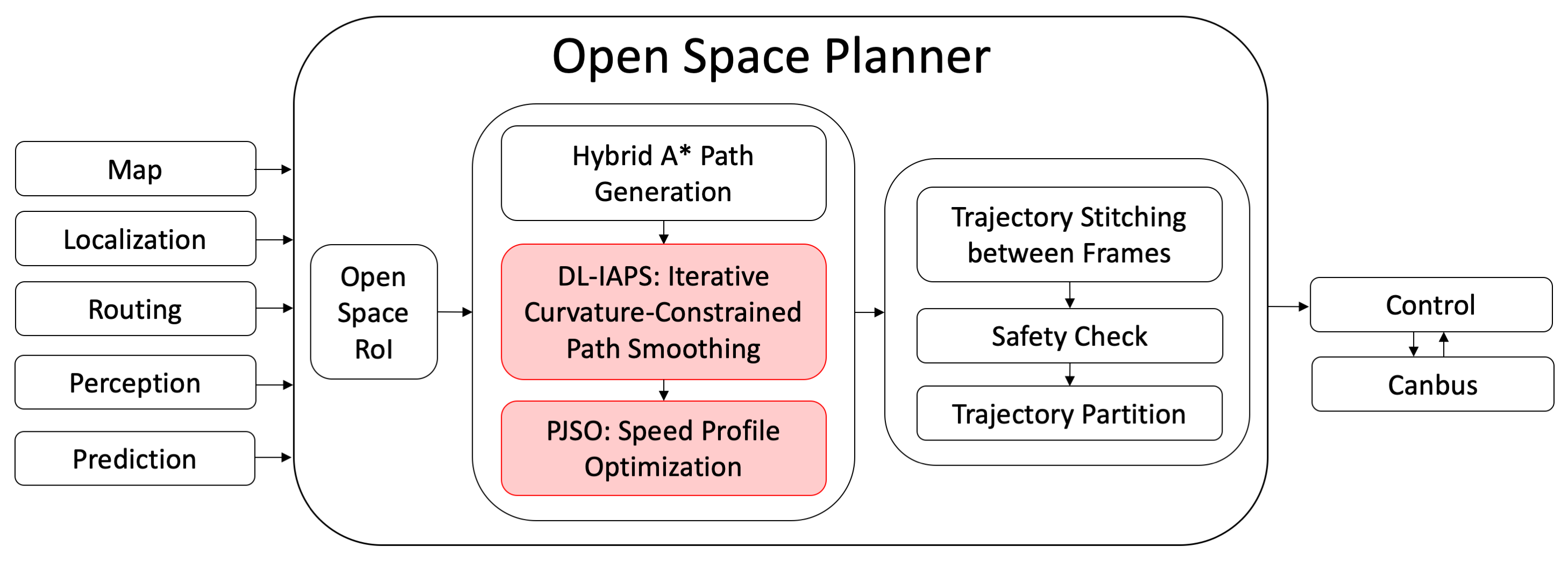}
\caption{Open Space Planner Architecture}
\label{figure:open_space_architect}
\end{figure}

\begin{enumerate}
	\item \textbf{Region of Interest (RoI)}: This module receives information from map and  perception, filters out far away or fast-moving obstacles and defines a task specific end position and collision free area for later modules
	\item \textbf{Trajectory Generation}: This module contains three parts: collision free Hybrid A* path searching, iterative curvature-constrained path smoothing (i.e., DL-IAPS) and speed profile optimization (i.e., PJSO).
	\item \textbf{Trajectory Post-processing}: This module contains three parts: trajectory stitching (with trajectory from last planning cycle), collision check (for fast-moving obstacles), and trajectory partition that splits trajectory into forward/backward pieces.
\end{enumerate}

As highlighted in Fig.~\ref{figure:open_space_architect}, our main focus in this paper is the design and implementation of the iterative curvature constrained path smoothing and speed profile optimization in the following chapters, assuming a collision free reference path $\mathcal{P} = \{P_i\}_{i=1}^n$ generated from upstream search based path planner (Hybrid A* in our case).

\section{Path Speed Decoupled Trajectory Optimization}
\label{sec:iterative_anchoring}
The path/speed decoupled trajectory planning contains two parts: in~\ref{subsec:path_smoothing} we introduce our path smoothing design and in~\ref{subsec:speed_profile_generation} we introduce our speed profile optimization.

\subsection{Dual-Loop Iterative Anchoring Path Smoothing}
\label{subsec:path_smoothing}
In this subsection, we introduce a Dual-Loop Iterative Anchoring Path Smoothing (DL-IAPS) for collision avoidance and path smoothing. The overall algorithm is shown in~\ref{algorithm:path_smoothing}. The inner loop starts from the collision free trajectory from Hybrid A* as reference path and smooth the path with curvature constraint via Sequential Convex Programming, and the outer loop check collision avoidance and shrink the corresponding state feasible region conditionally. Outer iteration terminates when the smoothed path passes collision check with all obstacles.

\begin{algorithm}
\caption{DL-IAPS path planning}
\label{algorithm:path_smoothing}
\SetAlgoLined
\SetKwInput{Variables}{Variables}
\SetKwInput{Parameters}{Parameters}

\Variables{
\begin{flushleft}
    \hspace{0.5em} $f:$ \text{cost function as in~\eqref{eq:adapted_cost}} \\
    \hspace{0.5em} $P_k, k = 0...n-1:$ \text{vehicle positions $[x_k, y_k]$}\\
    \hspace{0.5em} $g:$ \text{inequality constraint on curvature in~\eqref{eq:adapted_c5}} \\
    \hspace{0.5em} $s:$ \text{slack variable with respect to $g$} \\
    \hspace{0.5em} $t:$ \text{trust region size}\\
    \hspace{0.5em} $\mu:$ \text{penalty coefficient} \\
    \hspace{0.5em} $\mathcal{B}:$ \text{state bubble size}
\end{flushleft}
}
\Parameters{
\begin{flushleft}
    \hspace{0.5em} $\alpha:$ \text{penalty scaling factor} \\
    \hspace{0.5em} $\rho:$ \text{trust region adaptation threshold} \\
    \hspace{0.5em} $\gamma^+, \gamma^-:$ \text{trust region change ratio} \\
    \hspace{0.5em} $f_{tol}:$ \text{cost function convergence threshold} \\
    \hspace{0.5em} $x_{tol}:$ \text{decision variables convergence threshold} \\
    \hspace{0.5em} $c_{tol}:$ \text{constraint satisfaction threshold} \\
    \hspace{0.5em} $\beta:$ \text{state bubble change ratio}
\end{flushleft}
}
\algorithmiccomment{begin of outer loop for collision avoidance} \\
\For{Collision Check iteration = 1, 2, \dots} {
\algorithmiccomment{begin of inner loop for path smoothing} \\
    \For{Penalty iteration = 1, 2, \dots} { 
        \For{Sub-problem iteration = 1, 2, \dots} {
            $\hat{g} = linearization(g, P_{last-iteration})$ \\
            \For{Trust Region iteration = 1, 2, \dots} {
                $P \leftarrow \argmin_P f(P) + \mu \sum_{i = 0}^{m_{inequ}} s_i$ \\
                \eIf {TrueImprove/ModelImprove $> \rho$}{
                    $t \leftarrow t * \gamma^+$ ;\ \textbf{break}
                }{
                    $t \leftarrow t * \gamma^-$
                }
                \uIf{$t < x_{tol}$}{\textbf{break}}
            }
            \uIf{converged according to $x_{tol}$ or $f_{tol}$}{\textbf{break}}
        }
        \eIf{constraints satisfied to tolerance $c_{tol}$}{\textbf{break}}{
        $\mu \leftarrow \alpha * \mu$
        }
    } 
\algorithmiccomment{end of the inner loop} \\

    \For{Obstacle number = 1, 2, \dots}{
        \For{path point = 1, 2, \dots} { 
        \eIf{Full dimension collision detected}{
    $\mathcal{B}_k \leftarrow \beta * \mathcal{B}_k$ 
    }{
        \textbf{continue}
        }
        }
    }
} \algorithmiccomment{end of the outer loop}
\end{algorithm}

\subsubsection{Inner Loop for Curvature Constrained Path Smoothing}
\label{subsec:inner_loop}
 Inspired by some Elastic Band Approach path smoothing method~\cite{zhu2015convex}\cite{rosmann2015timed}, With vehicle turning radius $R$ and its minimum value $R_{min}$,  a relation in~\eqref{eq:original_curvature} between position vector change rate $A_k$ and maximum path curvature $1/R_{min}$ can be approximated based on the assumption that $P_k$ are uniformly and densely spread over the path ( $\norm{V_{k+1}} \approx \norm{V_{k}}$ based on the uniform distribution assumption, $sin(\theta_{k}) \approx \theta_{k}$ based on the small angle assumption due to the dense distribution, and $\theta_{k} \approx \norm{P_k - P_{k - 1}} / R $).
\begin{equation}
      \label{eq:original_curvature}
        \begin{aligned}
        & \norm{A_k} = \norm{V_{k+1} - V_{k}}
        \approx 2 * \norm{V_k} * sin(\theta_k /2) \\
        & \hspace{2.0em} \approx \norm{P_k - P_{k - 1}}^2 / R
          \leq \norm{P_k - P_{k - 1}}^2 / R_{min}
        \end{aligned} 
\end{equation}



It is worth mentioning that the work in CES deals with the
curvature constraint in~\eqref{eq:original_curvature} with an assumption that length of the smoothed path $\norm{P_k - P_{k-1}}^2$ in~\eqref{eq:adapted_c5} is about equal to that of the input reference path $\norm{P_k^{ref} - P_{k - 1}^{ref}}^2$ so that the order of curvature constraint can be reduced from quartic to quadratic~\cite{zhu2015convex}. However, the curvature performance comparison in Section~\ref{subsec:simple_simulation} shows CES's approach tends to invalidate the maximum curvature constraint when the smoothed path is much shorter than the reference path in extreme cases, which affects control feasibility.



With above path curvature constraint being a quartic constraint, the nonlinear path smoothing optimization problem is formulated in as:
\begin{subequations}
  \label{eq:adapted_original}
  \begin{align}
    \label{eq:adapted_cost}
    & \min_{\substack{%
        \text{$P$}
      }}\, 
      f(P) 
      = \min_{\substack{%
        \text{$P$}
      }}\,  
      \sum_{k=1}^{n - 2} \norm{A_k}^2 \\
    & \hspace{4.25em}= \min_{\substack{%
        \text{$P$}
      }}\,  
      \sum_{k=1}^{n - 2} \norm{2P_k - P_{k - 1} - P_{k + 1}}^2
      \\
    & \text{subject to:} \nonumber \\
    \label{eq:adapted_c1}
    & \hspace{2.0em} P_0 = P_{0_{ref}}, P_{n - 1} = P_{n-1_{ref}},  \\
    \label{eq:adapted_c2}
    & \hspace{2.0em} P_1 = P_{0_{ref}} + \norm{P_1 - P_0} * \hat{u}_{\phi_0}, \\
    \label{eq:adapted_c3}
    & \hspace{2.0em} P_{n - 2} = P_{n - 1_{ref}} + \norm{P_{n - 1} - P_{n - 2}} * \hat{u}_{\phi_{n - 1}}, \\
    \label{eq:adapted_c4}
    & \hspace{2.0em} P_k \in \mathcal{B}_k, \text{ for } k = 2, \dots n - 3, \\
    \label{eq:adapted_c5}
    & \hspace{2.0em} g(P) = \norm{2P_k - P_{k - 1} - P_{k + 1}}^2 - \dfrac{\norm{P_k - P_{k-1}}^4}{R_{min}^2} < 0 , \nonumber \\
    & \hspace{4.0em} \text{for } k = 1, \dots n - 2, 
  \end{align}
\end{subequations}

The notations in~\eqref{eq:adapted_original} are defined in Section~\ref{sec:problem_statement}. The optimization cost in~\eqref{eq:adapted_cost} tries to reduce the difference of one path point with respect to its neighboring point along the new trajectory. Such cost encourage every three consecutive points to be in a straight line therefore minimizing the curvature.
$\phi_0$ and $\phi_{n - 1}$ are headings of path's initial and end points respectively, which are same as reference path initial and end points headings.
$\hat{u}_{\phi}$ is the unit norm vector along the direction $\phi$.
$\mathcal{B}_k$ as shown in Fig.~\ref{figure:path_smoothing} is state bubble constraining the feasible region of a point's position in the optimization problem.

Equation~\eqref{eq:adapted_original} is hard to solve due to its non-linearity in constraints such as in~\eqref{eq:adapted_c4} and ~\eqref{eq:adapted_c5}, so we leverage SCP to solve it. SCP repeatedly approximate the original problem as a convex quadratic programming problem around current iteration point and solved it until convergence~\cite{schulman2014motion}. The related convex approximated sub-problem is then re-formulated as~\eqref{eq:adapted_convex}:
\begin{subequations}
  \label{eq:adapted_convex}
  \begin{align}
    \label{eq:convex_cost}
    & \min_{\substack{%
        \text{$P$, $d$}
      }}\,  
      \sum_{k=1}^{n - 2} \norm{2P_k - P_{k - 1} - P_{k + 1}}^2
      + \mu \sum_{k=1}^{n-1} s_k \\
    & \text{subject to:} \nonumber \\
    & \hspace{2.0em} \label{eq:convex_c1} 
        P_0 = P_{0_{ref}}, P_{n - 1} = P_{n-1_{ref}},  \\
    & \hspace{2.0em} \label{eq:convex_c2}
        P_1 = P_{0_{ref}} + \norm{P_1 - P_0} * \hat{u}_{\phi_0}, \\
    & \hspace{2.0em} \label{eq:convex_c3}
        P_{n - 2} = P_{n - 1_{ref}} + \norm{P_{n - 1} - P_{n - 2}} * \hat{u}_{\phi_{n - 1}}, \\
    & \hspace{2.0em} \label{eq:convex_c4}
        {Lx}_k \leq x_k \leq {Ux}_k, \text{ for } k = 2, \dots n - 3, \\
    & \hspace{2.0em} \label{eq:convex_c5}
        {Ly}_k \leq y_k \leq {Uy}_k, \text{ for } k = 2, \dots n - 3, \\
    & \hspace{2.0em} \label{eq:convex_trust_region_c1}
        {x}_k^{pre} - t \leq x_k \leq {x}_k^{pre} + t, \text{ for } k = 2, \dots n - 3, \\
    & \hspace{2.0em} \label{eq:convex_trust_region_c2}
        {y}_k^{pre} - t \leq y_k \leq {y}_k^{pre} + t, \text{ for } k = 2, \dots n - 3, \\
    & \hspace{2.0em} \label{eq:convex_c7}
        \hat{g}(P_k^{pre}, P_{k-1}^{pre}, P_{k+1}^{pre}, P_k, P_{k-1}, P_{k+1}) - s_k < 0, \\
    & \hspace{4.0em} \text{for } k = 1, \dots n - 1, \nonumber \\
    & \hspace{2.0em} s_k \geq 0, \text{ for } k = 1, \dots, n - 2. 
  \end{align}
\end{subequations}

State feasible bubble $\mathcal{B}_k$ constraints~\eqref{eq:adapted_c4} is approximated as an inscribed box with ${Ux}_k, {Uy}_k$, the upper and ${Lx}_k, {Ly}_k$, the lower limit constraints~\eqref{eq:convex_c4}\eqref{eq:convex_c5}. Trust region state constraints respect to previous iteration is shown in constraints~\eqref{eq:convex_trust_region_c1}\eqref{eq:convex_trust_region_c2} Nonlinear constraints~\eqref{eq:adapted_c5} are transformed to linearized constraints~\eqref{eq:convex_c7} around previous iteration path points $P^{pre}_k$ via Euler method.
Trust Region method~\cite{HUANG2015107} is then applied afterward to guarantee the approximation quality between steps:  TrueImprove/ModelImprove in algorithm~\ref{algorithm:path_smoothing} is the ratio between true improvement to objective and constraint violation of original problem~\ref{eq:original_curvature} to those of the sub-problem~\cite{schulman2014motion}. We enlarge or shrink trust region size for each sub-problem according to this ratio.

\subsubsection{Outer Loop for Collision Avoidance}
\label{subsec:outer_loop}

\begin{figure}[!h]
 \centering
\includegraphics[trim=0 0 0 0, clip, width=0.85\linewidth]{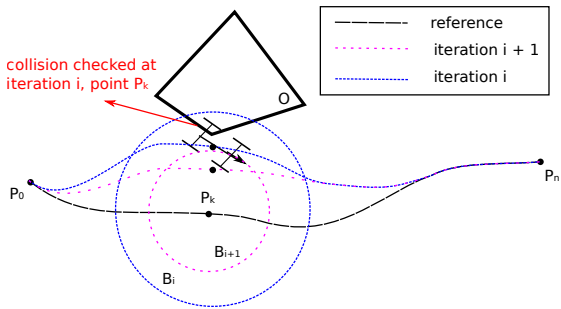}
 \caption{Illustration of collision check and $\mathcal{B}_k$ updates}
 \label{figure:path_smoothing}
\end{figure}

Following the path smoothing in the inner loop, we check whether the generated path trajectory collides with obstacles. The precise shapes of obstacles and the ego vehicle are considered during the collision check.
If collision is detected with $k$th path point , we shrink the corresponding bubble $\mathcal{B}_k$ size by a ratio $\beta < 1$.
The detailed procedure is illustrated in Fig.~\ref{figure:path_smoothing}.

Although some prior works~\cite{zhu2015convex}\cite{chen_liu_tomizuka_2018} avoid using precise collision check to speed up the trajectory generation process, we found it essential to ensure vehicle’s safely operation, especially in some scenarios with narrow spaces, such as pull over and parallel parking.

It is also worthy noted that, instead of directly anchoring the point back to related reference points as in ~\cite{dolgov2010path}, we shrink the state space around the collision path point iteratively, with the purpose of avoiding over-sacrifice the path smoothness (which is critical in Robotaxi operations) due to collision avoidance.

\subsection{Piece-wise Jerk Speed Optimization}
\label{subsec:speed_profile_generation}

In this subsection, we introduce the Piece-wise Jerk Speed Optimization (PJSO) method to generate longitudinal speed profile along the path generated from Subsection~\ref{subsec:path_smoothing}. As the path generated by the DL-IAPS oftentimes comprises both forward and back vehicle movement, the speed optimization is done separately on each piece assuming the vehicle always comes to a complete stop at the gear shifting position for better driving comfort.

We treat the speed profile optimization problem as longitudinal traversal distance smoothing along a time horizon $T_{horizon}$ discretized by $\Delta{t}$. The decision variables includes $[s_k, \dot{s}_k, \ddot{s}_k]$ for $k = i, \dots, n-1$, where $n = T_{horizon}/\Delta{t}$, and $s_k, \dot{s}_k, \ddot{s}_k$ are the longitudinal traversal distance, speed and acceleration. We use a cubic polynomial as the state dynamics between $[s_k, \dot{s}_k, \ddot{s}_k]$ and $[s_{k+1}, \dot{s}_{k+1}, \ddot{s}_{k+1}]$, assuming the jerk (i.e.,rate of change of acceleration) is constant from time $t_k$ to $t_{k+1}$ (which is so-called "piece-wise" jerk). The dynamics are shown in ~\eqref{eq:speed_dynamics}:
\begin{subequations}
  \label{eq:speed_dynamics}
  \begin{align}
        & \dot{s}_{k+1} = \dot{s}_{k} + \ddot{s}_{k} \Delta{t} + \frac{1}{2} \dddot{s}_{k, k + 1} \Delta{t}^2 \nonumber \\
        & \hspace{2.0em}
        = \dot{s}_{k} + \frac{1}{2} \ddot{s}_{k} \Delta{t} + \frac{1}{2} \ddot{s}_{k + 1} \Delta{t},  \\
        & s_{k+1}
        = s_k + \dot{s}_{k} \Delta{t} + \frac{1}{2} \ddot{s}_{k} \Delta{t}^2 + \frac{1}{6} \dddot{s}_{k, k + 1} \Delta{t}^3 \nonumber \\
        & \hspace{2.0em}
        = s_k + \dot{s}_{k} \Delta{t} + \frac{1}{3} \ddot{s}_{k} \Delta{t}^2 + \frac{1}{6} \ddot{s}_{k + 1} \Delta{t}^2.
  \end{align}
\end{subequations}

Such problem formulation makes it flexible to set constraints for feasibility. $\dot{s}$, $\ddot{s}$ and $\dddot{s}$ are set to be constrained by vehicle parameters $\mathcal{S} \subset \R^4$. The curvature-induced speed constraint on $\dot{s}$ can be added to the problem as~\eqref{eq:s_dot_curvature_constraint} with the actual path curvature function $\kappa(s)$, but to keep it a quadratic programming form, we approximate the speed constraint induced by path curvature as a linear constraint in~\eqref{eq:s_dot_constraint}, with maximum lateral acceleration~$lateral\_a_{max}$ and maximum curvature ~$\kappa(s)_{max}$ along the generated path. It would over limit the speed where the path curvature is not at its maximum but still be a proper constraint as the problem setting of the algorithm is not racing competition but relatively slow free space maneuvering, like parallel parking.  
\begin{equation}
 \label{eq:s_dot_curvature_constraint}
 \hspace{2.0em} \dot{s}_j < \sqrt{a_{lateral\_max} / \kappa(s_j)}, \hspace{0.5em} \text{for } j = 0, \dots, n-1
\end{equation}

With the optimization constraints been set up, the optimization step horizon $n$, which decides the time horizon by $T_{horizon} = n\Delta{t}$, is initialized in~\eqref{eq:n_constraint}. As $n$ can't be too short to traverse through the entire path, we first estimate its feasible lower bound $n_{min}$ based on the vehicle dynamics. Given the maximum acceleration $a_{max}$, maximum speed $v_{max}$ and the total traverse path distance $s_{f}$, we have $n_{min} = \frac{v_{max}^2 + s_{f} a_{max}}{a_{max} v_{max} \Delta{t}}$, with an infinite jerk assumption so that the vehicle is able to accelerate by $a_{max}$ to peak speed $v_{max}$ and decelerates by $-a_{max}$ to zero speed. Then, with a constrained jerk, the horizon is multiplied by a heuristic expansion ratio $r$ as $n = r*n_{min}$, where $r$ is selected in range of $[1.2, 1.5]$. Higher ratio gives more dynamic feasibility for this fixed-distance speed optimization, but an over-estimated $r$ may bring in unnecessary computation time as it increases the dimension of decision variables.

To minimize the traversal time to path end at $s_f$, we set a cost term $ \sum_{k=0}^{n-1}(s_k - s_{f})^2$ in objective to penalize the distance gap between every-step state and final state. In addition to that, to balance the driving comfort, penalties on $\ddot{s}$ and $\dddot{s}$ are included.

The complete quadratic programming optimization formulation with weighting hyperparameter $w_{s_f}$, $w_{\dddot{s}}$ and $w_{\ddot{s}}$ is presented in~\eqref{eq:s_penalty2} as:
\begin{subequations}
  \label{eq:s_penalty2}
  \begin{align}
      & \min_{\substack{%
        \textbf{$s$, $\dot{s}$, $\ddot{s}$}
      }}\,
      \mathcal{J}_c~\pb{\text{$s$, $\dot{s}$, $\ddot{s}$}}
              = w_{s_f} \sum_{k=0}^{n-1}(s_k - s_{f})^2 \nonumber \\ 
              & \hspace{2.0em} + w_{\dddot{s}} \sum_{k=0}^{n-2} ((\ddot{s}_{k + 1} - \ddot{s}_{k}) / \Delta{t})^2  
              + w_{\ddot{s}} \sum_{k=0}^{n-1} \ddot{s}_k^2, \label{eq:s_cost}\\
      & \text{subject to:} \nonumber\\
              & \hspace{2.0em} [s_0, \dot{s}_0, \ddot{s}_0] = [0.0, 0.0, 0.0], \\
              & \hspace{2.0em} [s_j, \dot{s}_j, \ddot{s}_j, \frac{\ddot{s}_{j + 1} - \ddot{s}_{j}}{\Delta{t}}] \in \mathcal{S}, \\
              & \hspace{2.0em} \dot{s}_j < \sqrt{a_{lateral\_max} / \kappa(s)_{max}}, \label{eq:s_dot_constraint} \\
              & \hspace{2.0em}
              \dot{s}_{k+1}
                  = \dot{s}_{k} + \frac{1}{2} \ddot{s}_{k} \Delta{t} + \frac{1}{2} \ddot{s}_{k + 1} \Delta{t}, \label{eq:s_d1} \\
              & \hspace{2.0em}
                  s_{k+1}
                  = s_k + \dot{s}_{k} \Delta{t} + \frac{1}{3} \ddot{s}_{k} \Delta{t}^2 + \frac{1}{6} \ddot{s}_{k + 1} \Delta{t}^2, \label{eq:s_d2} \\
              & \hspace{3.0em} \text{for } k = 0, \dots, n-2, ~j = 0, \dots, n-1,  \nonumber \\
              & \hspace{4.5em} \text{and } n = r \frac{v_{max}^2 + s_{f} a_{max}}{a_{max} v_{max} \Delta{t}}. \label{eq:n_constraint} \\
              \nonumber
  \end{align}
\end{subequations}


\section{experiment results and applications on the apollo platform}
\label{sec:experiments}
In this section, we present both numerical simulations and real-world vehicle testing results on Apollo Open Source Autonomous Driving Platform, to demonstrate control feasibility, computation efficiency and robustness of  proposed optimization methods.

\subsection{Numerical Simulations: Performance Validation on Control Feasibility and Computation Efficiency}
\label{subsec:simple_simulation}

For the autonomous driving application, the control feasibility (i.e., path smoothness and physical constraints satisfaction) and the computation efficiency, are two crucial metrics to evaluate the performance of a planning optimization method. To validate that our proposed DL-IAPS plus PJSO optimization algorithm actually reaches a good balance of control feasibility among the common autonomous driving planning methods, we evaluate our planner with batch simulation tests via a standard parallel parking scenario and compare its performance with aforementioned H-OBCA~\cite{zhang2018autonomous} which is a path speed coupled trajectory optimization and CES path planners ~\cite{zhu2015convex}. 

\begin{figure}[ht]
 \centering
\includegraphics[width=1.1\linewidth]{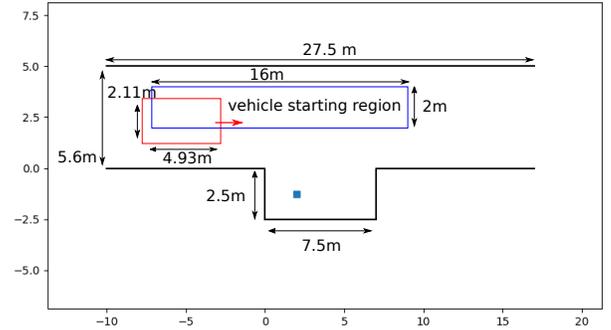}
 \caption{Illustration of standard parallel parking simulation environment with test set-ups: vehicle wheelbase: 2.8 $m$;
 path curvature ($m^{-1}$): $[-0.2, 0.2]$;
 speed ($m/s$): $[-1, 2]$;
 acceleration ($m/s^2$): $[-1, 1]$;
 acceleration change rate($m/s^3$): $[-1, 1]$;
 Hybrid A* step size(m): 0.2;
 Hybrid A* steering resolution(rad): 0.026;
 Path smoothing $\Delta{s}$(m): 0.1;
 Speed profile optimization $\Delta{t}$(s): 0.05.
 }
 \label{figure:simulation_scenario}
\end{figure}

\begin{figure}[ht]
 \centering
\includegraphics[width=0.99\linewidth]{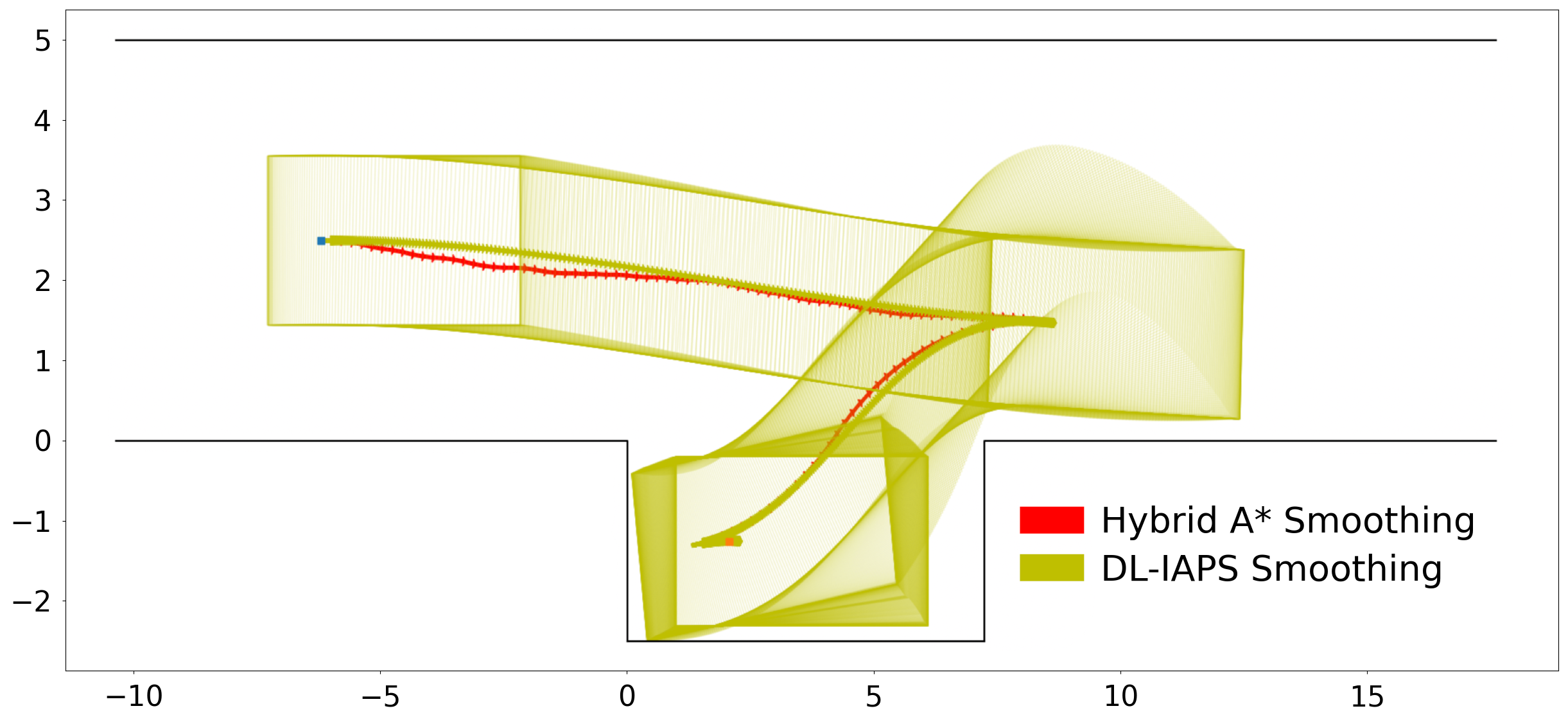}
 \caption{Optimized path trajectory from starting pose [x = -6m, y = 2.5m, $\theta$ = 0.0] with Hybrid A* path generator and DL-IAPS smoothing}
 \label{figure:parallel_parking}
\end{figure}

First, a standard numeric testing environment of parallel parking scenario is set up as Fig.~\ref{figure:simulation_scenario}. Parallel parking and another similar pull over scenarios (which have almost the same trajectory planning and only differ from how to formulate the RoI) , integrate with all kinds of complex vehicle behaviors, including the large-scale pose adjustment in a narrow space with irregular obstacles/boundaries, multiple forward/backward driving switching, and potential multiple obstacles and complex environmental boundaries. Therefore, parallel parking (or pull over) scenario is usually utilized as a typical test case to evaluate the path smoothness, control feasibility and computation efficiency of the free space planner. With a fixed ending parking pose, 80 different starting poses are tested in the simulation, by gridding the configuration space within $x \in [-8, 8]\ m$ with interval $1.0\ m$ and $y \in [2, 4]\ m$ with interval $0.5\ m$ with zero heading angle $\theta$. The proposed algorithms are implemented on Apollo Platform, and simulated in an environment with an i7 processor clocked at 2.6 GHz. The quadratic programming problem in both path smoothing and speed optimization are solved by a QP solver, OSQP~\cite{osqp}.

\begin{figure}[ht]
 \centering
\includegraphics[width=0.99\linewidth]{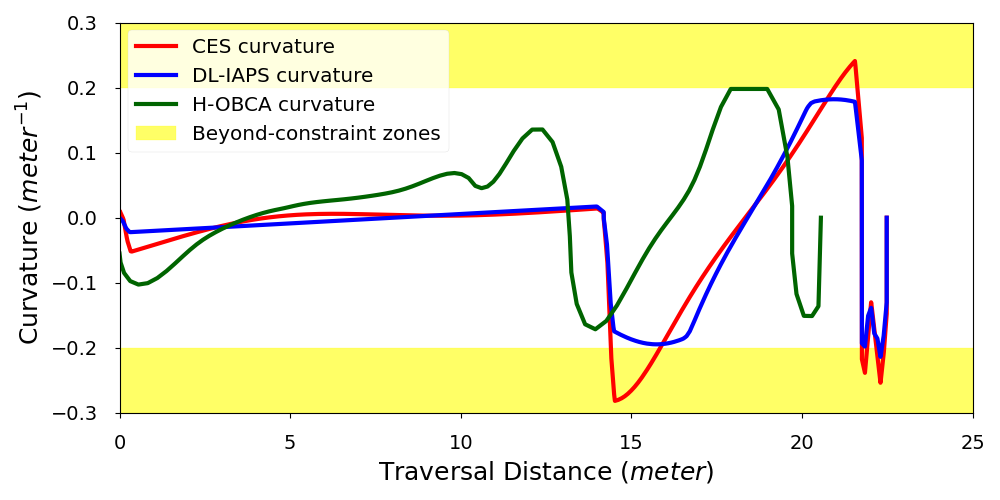}
 \caption{Optimized path (curvature) trajectories comparison among the DL-IAPS, CES and H-OBCA optimizations}
 \label{figure:kappa_compare}
\end{figure}

\begin{figure}[ht]
 \centering
\includegraphics[width=0.99\linewidth]{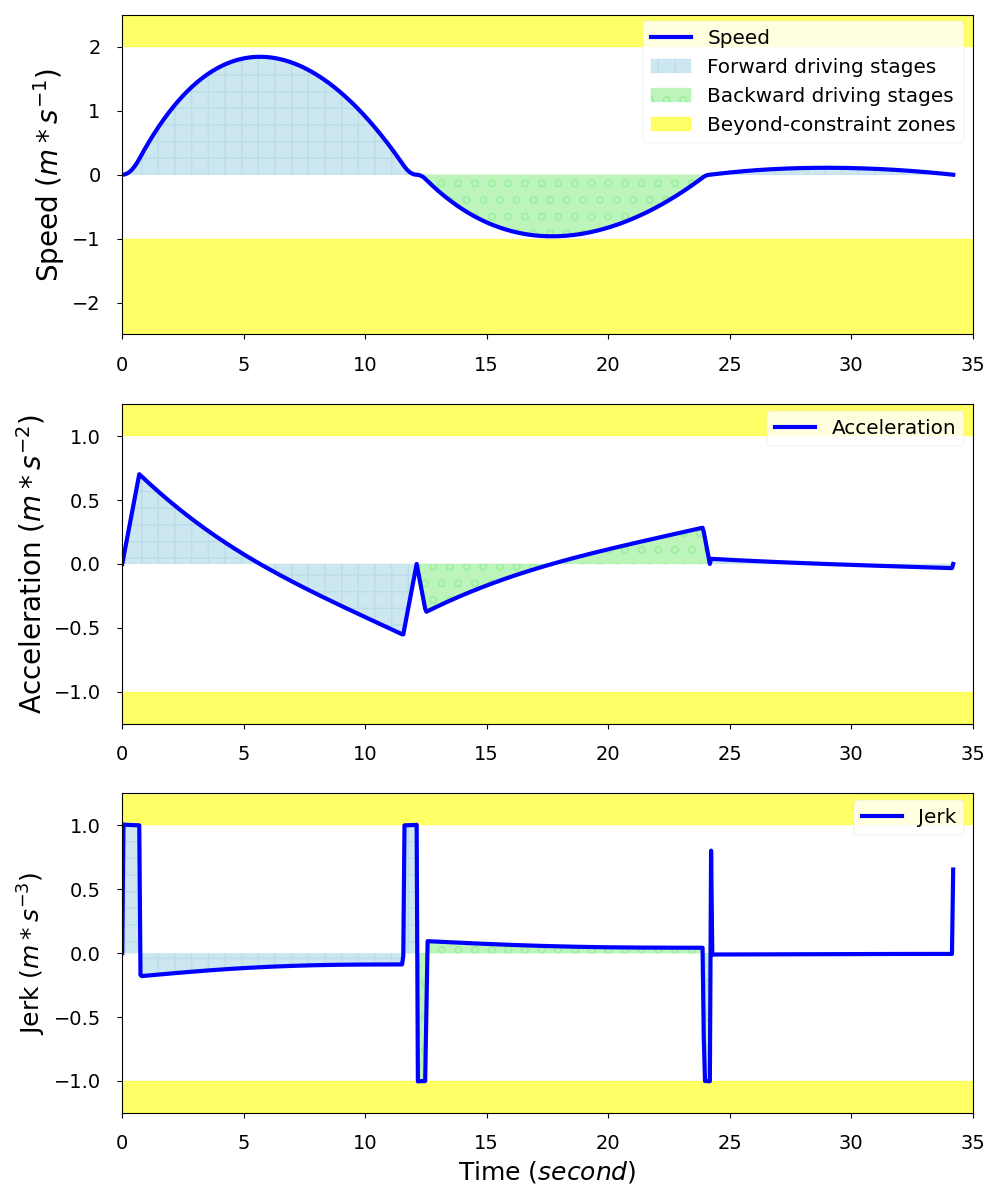}
 \caption{Optimized speed, acceleration and jerk trajectories}
 \label{figure:speed_profile}
\end{figure}

From the aforementioned simulation test environment, we implement our DL-IAPS plus PJSO planner together with the H-OBCA and CES to demonstrate their control feasibility and computation efficiency respectively, as follows:

\subsubsection{Control Feasibility (Smoothness and Curvature Constraints)}
\label{sec:curvature_constraint_validation_test}

First, the smoothness and constraint satisfaction of our proposed DL-IAPS optimized path trajectory are demonstrated and compared with different benchmarks in Fig.~\ref{figure:parallel_parking} and Fig.~\ref{figure:kappa_compare}. From Fig.~\ref{figure:parallel_parking}, the DL-IAPS optimized path is obviously more smooth than the one generated by the basic Hybrid A* algorithm; the latter is jerky because of its discretization of state space. More significantly, Fig.~\ref{figure:kappa_compare} demonstrates the control-feasible performance comparisons of our proposed planner and other two algorithms, where the yellow areas denote the forbidden zones in which the path curvatures are beyond the control-feasible and physical-realizable thresholds. With our DL-IAPS planner, the optimized path curvatures are well constrained by the maximum curvature (which is, the reciprocal of the minimal vehicle turning radius) decided by~\eqref{eq:adapted_original}, even at some extreme instances where the large path curvatures are needed (i.e., the $100\%$ full steering needs to be executed); while, with the CES algorithms, at these extreme instances the curvature actually exceeds the control-feasible constraints and results in the failed trajectory tracking, because~\eqref{eq:adapted_c5} is replaced by the approximation method from CES~\cite{zhu2015convex}. The H-OBCA algorithm presents the similar smoothness and constraint satisfaction with our DL-IAPS and however, the relatively lower computation efficiency (which will be proven in the next sub-session). 

Further, to better demonstrate the effectiveness of path smoothing and curvature constraints in our algorithm, the PJSO generated speed, acceleration and jerk (acceleration change rate) trajectories are shown in Fig.~\ref{figure:speed_profile}, in which the different shallow zones describe separate driving stages with either forward or backward gear of the vehicle. It can been seen that the speed/acceleration/jerk optimized by PJSO well balances driving comfort and minimal traversal time. 

\subsubsection{Computation Efficiency}
\label{sec:computation_time_test}

Although both the H-OBCA and our DL-IAPS plus PJSO present similar smoothness and control feasibility, the batch simulation results (consisting of 80 tests with different stating poses) demonstrate the better computation efficiency with our algorithm, as shown in Table~\ref{table:simulations}. The total time with our two-step path smoothing and speed optimization is only around 70ms in average, which is acceptable to most real-time applications; however, with highly similar simulation setups, the H-OBCA, which integrates path/speed smoothing and obstacle avoidance in just one-step NMPC-based trajectory planning~\cite{zhang2018autonomous}, asks for more than 1240ms running time, which is one order of magnitude more than our decoupling-based planner. 

\begin{table}[ht]
\centering
\caption{Computation time (in average through 80 cases with different starting poses), in (s). Reference path is generated via hybrid A* with extra average time cost of 0.4s}
\setlength{\tabcolsep}{0.5em}
\normalsize
\begin{tabular}{ p{4.25cm} | p{.8cm} p{.8cm} p{.8cm} }
 \hline
 Modules & mean & min & max\\ 
 \hhline{=:===}
 \textbf{DL-IAPS path smoothing} & \textbf{0.035}& 0.002 & 0.082\\
 \hline
 \textbf{PJSO speed optimization} & \textbf{0.035} & 0.021 & 0.070\\
 \hline 
 \textbf{Path Speed Total} & \textbf{0.07} & 0.023 & 0.152 \\
 \hhline{=:===}
 \textbf{H-OBCA Total} & \textbf{1.247} & 0.313 & 4.019 \\
 \hline
\end{tabular}
\label{table:simulations}
\end{table}

\subsection{Numerical Simulations: Expand Performance Validation with Complex Boundaries and Obstacles}
\label{sec:computation_time_boundaries_obstacles}

\begin{figure}[ht]
\begin{minipage}{0.16\textwidth}
 \centering
 \subfloat[Parking: 3 static obstacles \label{figure:simulation_scenario_obstacles_1}]{\includegraphics[width=.9\linewidth]{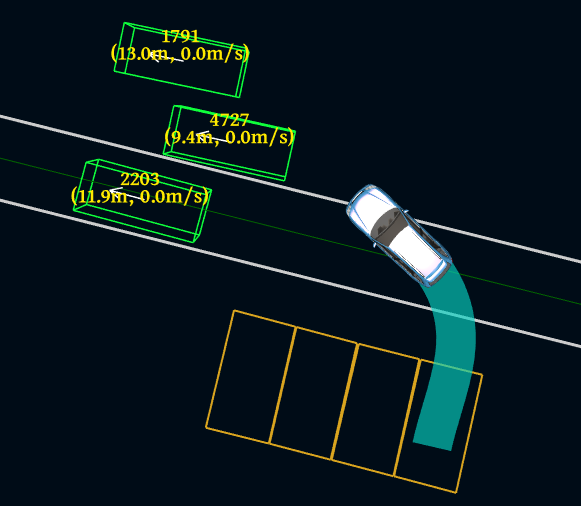}}
\end{minipage}\hfill
\begin{minipage}{0.16\textwidth}
 \centering
 \subfloat[Parking: 2 moving pedestrians \label{figure:simulation_scenario_obstacles_3}]{\includegraphics[width=.9\linewidth]{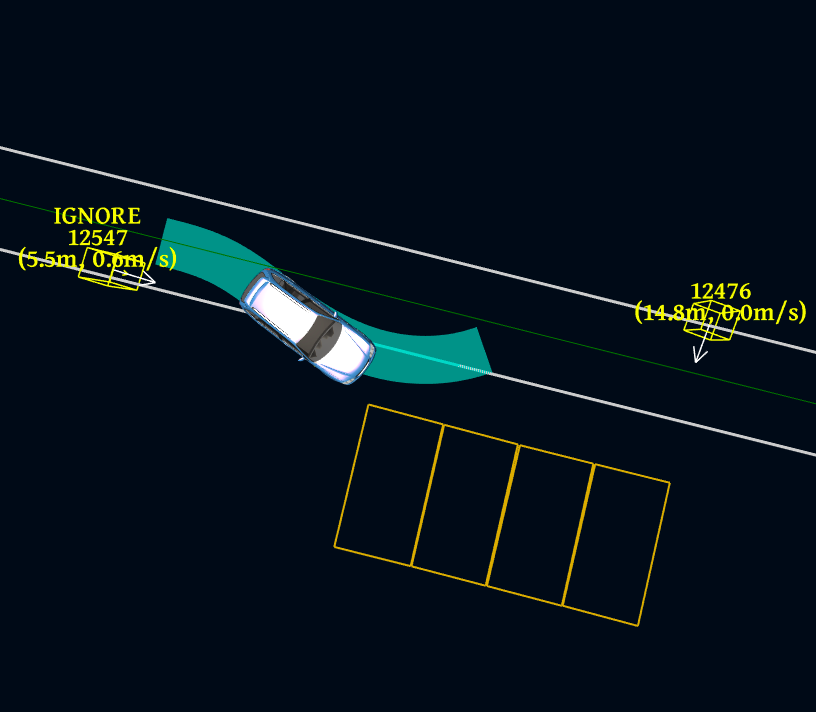}}
\end{minipage}\hfill
\begin{minipage}{0.16\textwidth}
 \centering
 \subfloat[Parking: 2 static obstacles \label{figure:simulation_scenario_obstacles_4}]{\includegraphics[width=.9\linewidth]{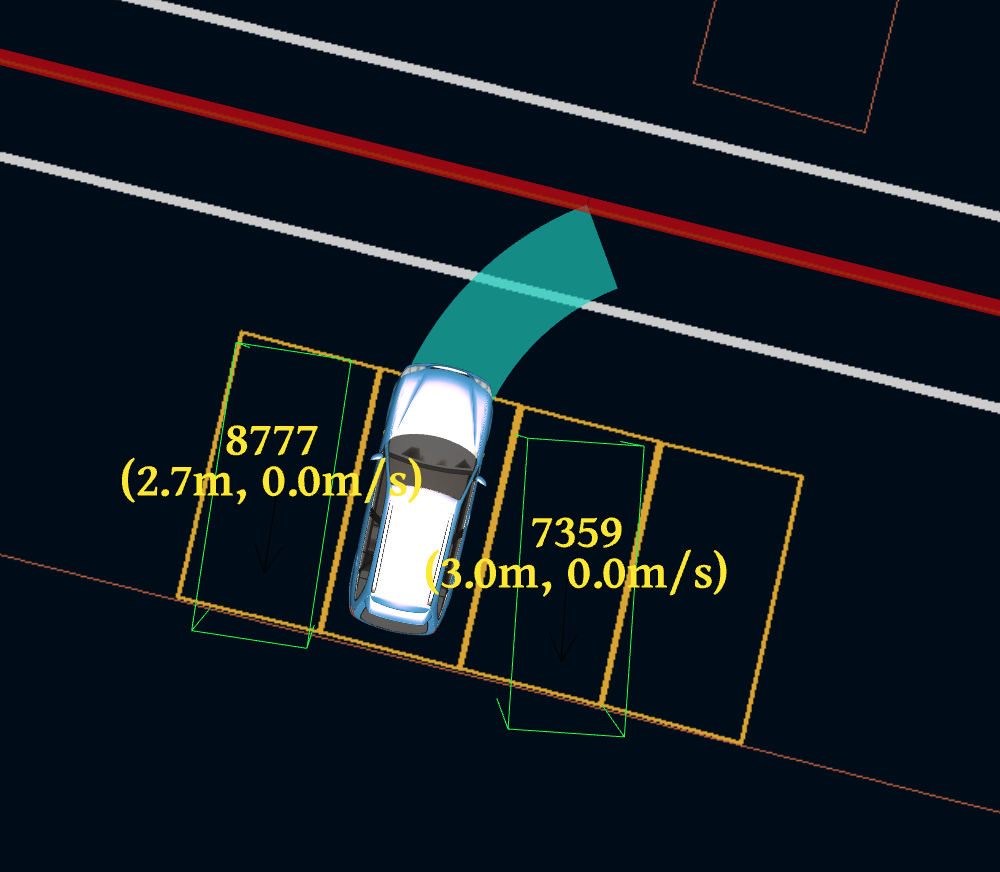}}
\end{minipage}\hfill
\begin{minipage}{0.16\textwidth}
 \centering
 \subfloat[Parking: 1 moving (out) obstacle \label{figure:simulation_scenario_obstacles_5}]{\includegraphics[width=.9\linewidth]{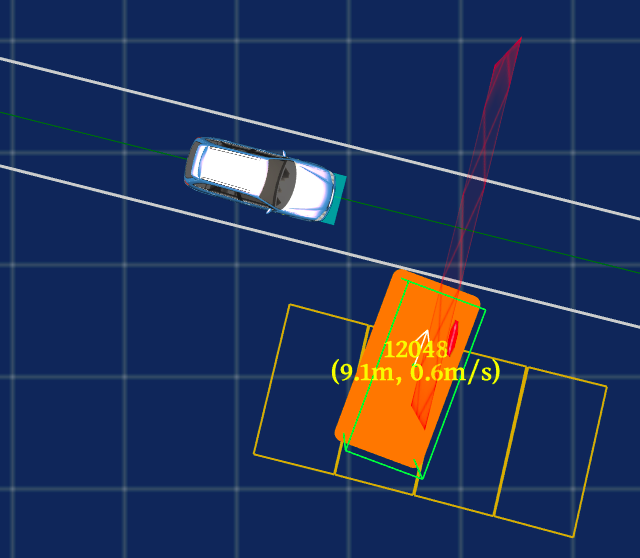}}
\end{minipage}\hfill
\begin{minipage}{0.16\textwidth}
 \centering
 \subfloat[Pull over: 5 static obstacles \label{figure:simulation_scenario_obstacles_2}]{\includegraphics[width=.9\linewidth]{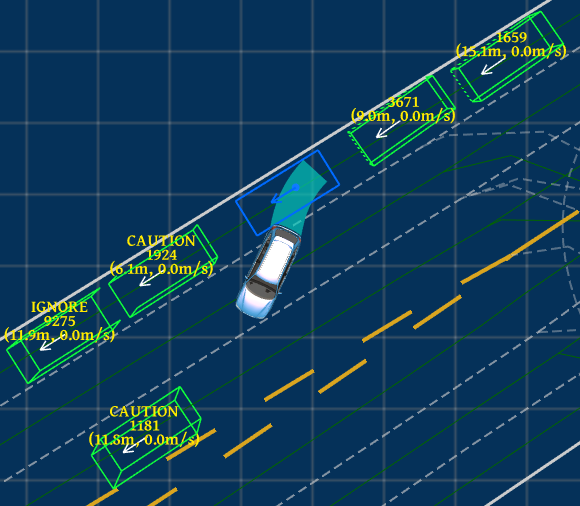}}
\end{minipage}\hfill
\begin{minipage}{0.16\textwidth}
 \centering
 \subfloat[Pull over: 4 static obstacles \label{figure:simulation_scenario_obstacles_6}]{\includegraphics[width=.9\linewidth]{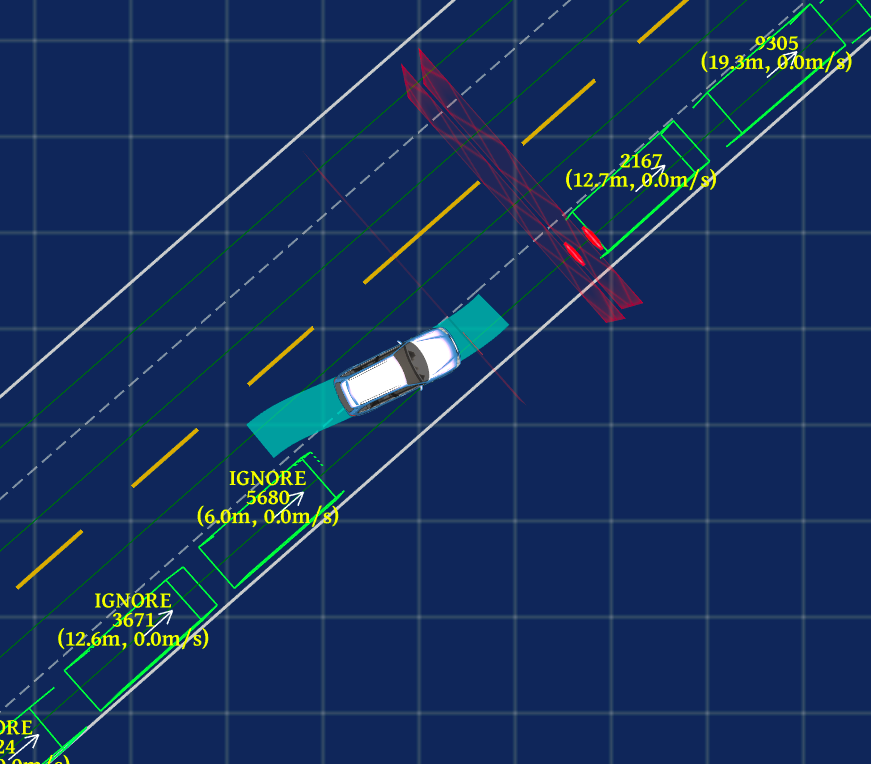}}
\end{minipage}\hfill
\caption{Various simulation test cases with multiple numbers of boundaries and obstacles (with same set-up parameters as in Figure~\ref{figure:simulation_scenario} except speed profile optimization $\Delta{t}$(s): 0.5)}
\label{figure:visualization_boundaries_obstacles}
\end{figure}

To scale the computation efficiency and validate the robustness to cope with complex obstacles and boundaries, we perform a large-scale, end-to-end simulation on Apollo online simulation platform that carries out totally 208 different free space test scenarios.
Fig.~\ref{figure:visualization_boundaries_obstacles} shows some of these free space test cases. With the purpose of identifying the sensitivity of optimization time consumption to the amount of the boundaries/obstacles, multiple obstacles are intentionally inserted into typical test cases. 

Table~\ref{table:scaling_boundaries_obstacles} demonstrates that for typical valet parking and pull over scenarios, the total computation time including path smoothing and speed optimization only slightly increases as the numbers of boundaries and obstacles increases and therefore, prevent the computation time from explosively growing induced by extensive obstacles or serpentine boundaries. 

\begin{table}[ht]
\centering
\caption{Computation time of valet parking and pull over test cases with multiple boundaries and obstacles, in (s)}
\setlength{\tabcolsep}{0.5em}
\begin{tabular}{p{0.9cm} | p{1.3cm} | p{0.84cm} | p{0.84cm}| p{0.84cm}| p{0.84cm} | p{0.90cm}}
 \hline
 Cases & Number of (Boundary, Obstacle) & Path Smoothing & Speed Profile & Total Time & Smooth Points & Time per Point \\ 
 \hhline{=:=:=:=:=:=:=}
  & (6 , 0) & 0.087 & 0.096 & \textbf{0.183} & 162 & \textbf{0.001130}\\
 \hhline{~|-|-|-|-|-|-}
  Parking & (6 , 1) & 0.107 & 0.081 & \textbf{0.188} & 162 & \textbf{0.001160}\\
 \hhline{~|-|-|-|-|-|-}
  & (6 , 2) & 0.106 & 0.082 & \textbf{0.188} & 162 & \textbf{0.001160}\\
 \hhline{~|-|-|-|-|-|-}
  & (6 , 3) & 0.106 & 0.078 & \textbf{0.184} & 162 & \textbf{0.001136}\\
 \hhline{=:=:=:=:=:=:=}
  & (4 , 2) & 0.104 & 0.100 & \textbf{0.204} & 258 & \textbf{0.000791} \\
 \hhline{~|-|-|-|-|-|-}
  Pull & (4 , 3) & 0.103 & 0.098 & \textbf{0.201} & 253 & \textbf{0.000794}\\
 \hhline{~|-|-|-|-|-|-}
  Over& (4 , 4) & 0.103 & 0.099 & \textbf{0.202} & 253 & \textbf{0.000798}\\
 \hhline{~|-|-|-|-|-|-}
  & (4 , 5) & 0.101 & 0.105 & \textbf{0.206} & 253 & \textbf{0.000814}\\
 \hline
\end{tabular}
\label{table:scaling_boundaries_obstacles}
\end{table}


\subsection{On-Road Experimental Implementation and Results}
\label{subsec:road_test}

To further demonstrate the control feasibility and real-world executability of the proposed DL-IAPS plus PJSO trajectory optimization algorithm, the planner is embedded in the the Apollo Autonomous Driving Platform and implemented in the real autonomous vehicles. 

\begin{figure}[ht]
\centering
\includegraphics[width=1.0\linewidth]{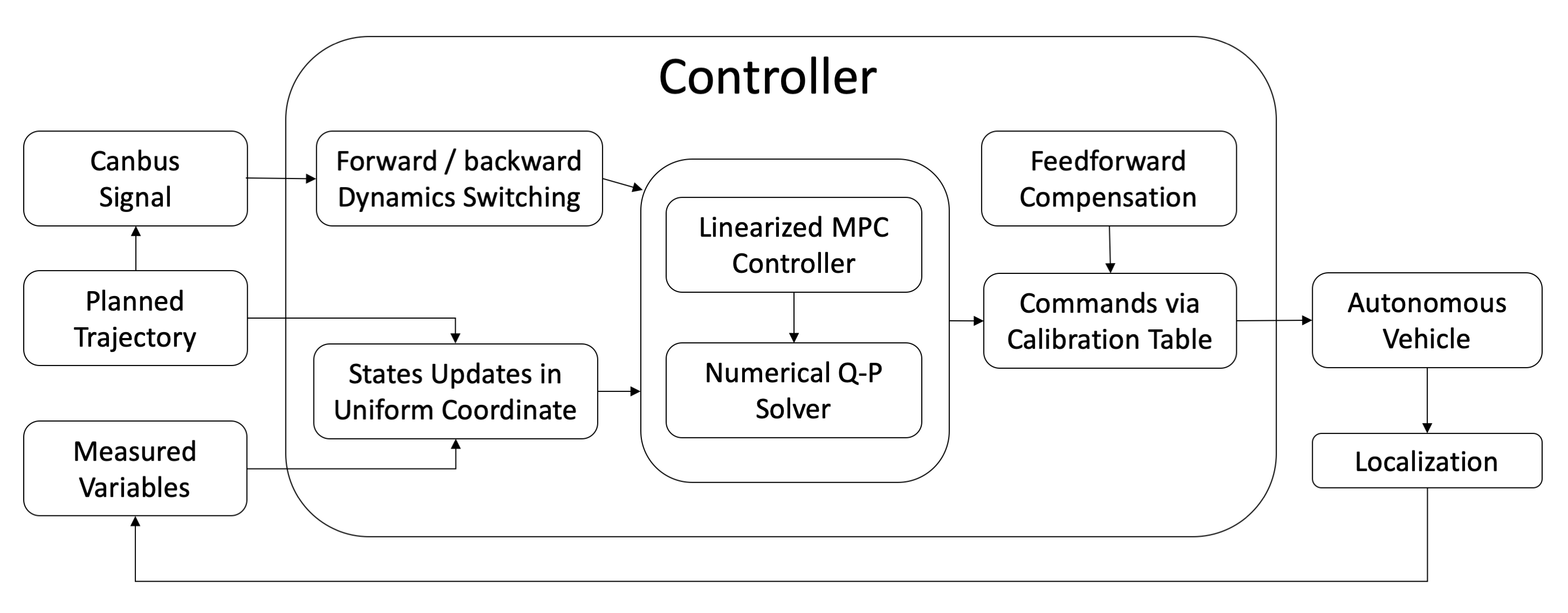}
\caption{On-road test vehicle controller architecture}
\label{figure:open_space_controller}
\end{figure}

As a lower-level executor to the planner module, the vehicle controller in the autonomous driving platform, as shown in Fig. ~\ref{figure:open_space_controller}, cooperates with the proposed free space planner to realize the optimized planning trajectory. The vehicle controller architecture contains three main components: error states generator, linearized Model Predictive Controller (MPC) and related Quadratic Programming (QP) solver, and feedforward control mapping (i.e., calibration table). The control performance in the free space scenarios highly depends on the smoothness and constraint satisfaction of the free space planning trajectory.

\begin{figure}[ht]
\begin{minipage}{0.16\textwidth}
 \centering
 \subfloat[US Field Test \label{figure:Pullover_Picture}]{\includegraphics[width=0.95 \linewidth, trim=0 0 0 35, clip]{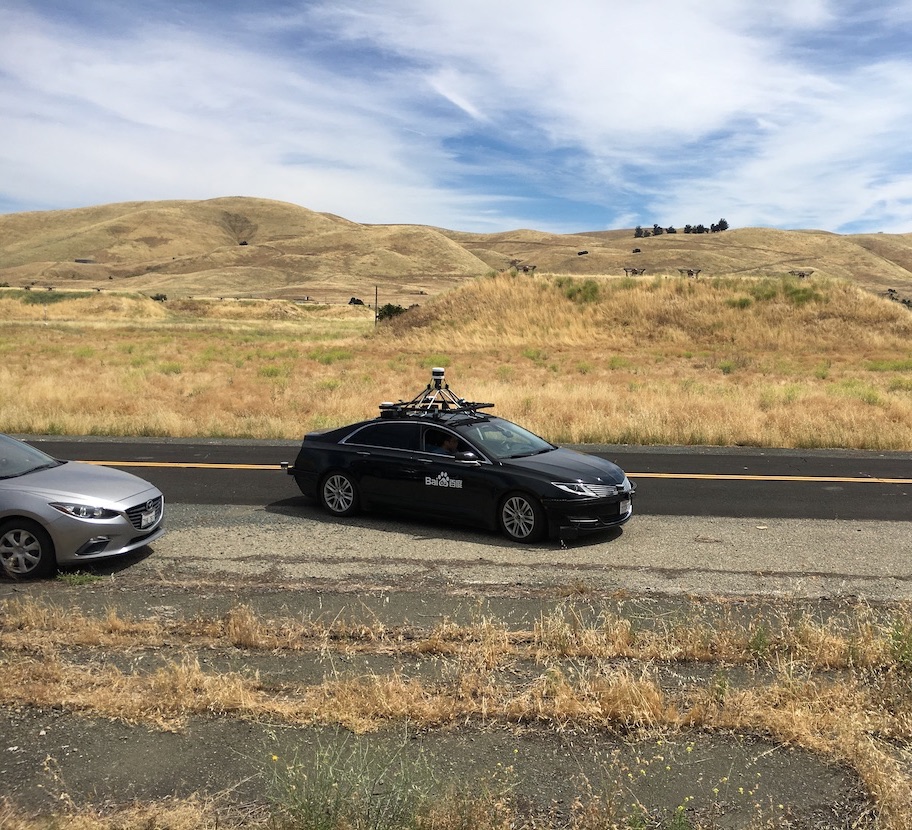}}
\end{minipage}\hfill
\begin{minipage}{0.16\textwidth}
 \centering
 \subfloat[Planning Trajectory\label{figure:Pullover_Trajectory_Optimizer_Visualization}]{\includegraphics[width=.95\linewidth]{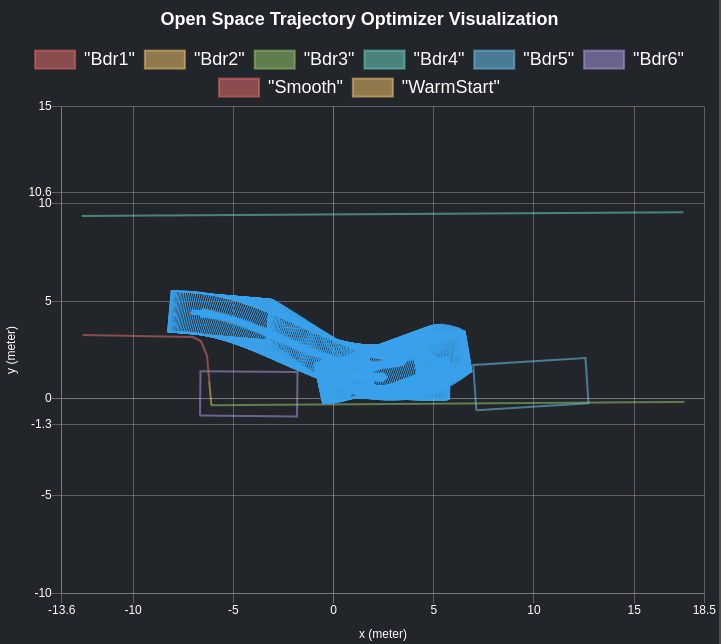}}
 \end{minipage}\hfill
\begin {minipage}{0.16\textwidth}
 \centering
 \subfloat[Test Data Visual \label{figure:Pullover_Stage2_Retry_Approach_Parking_1}]{\includegraphics[width=0.93\linewidth]{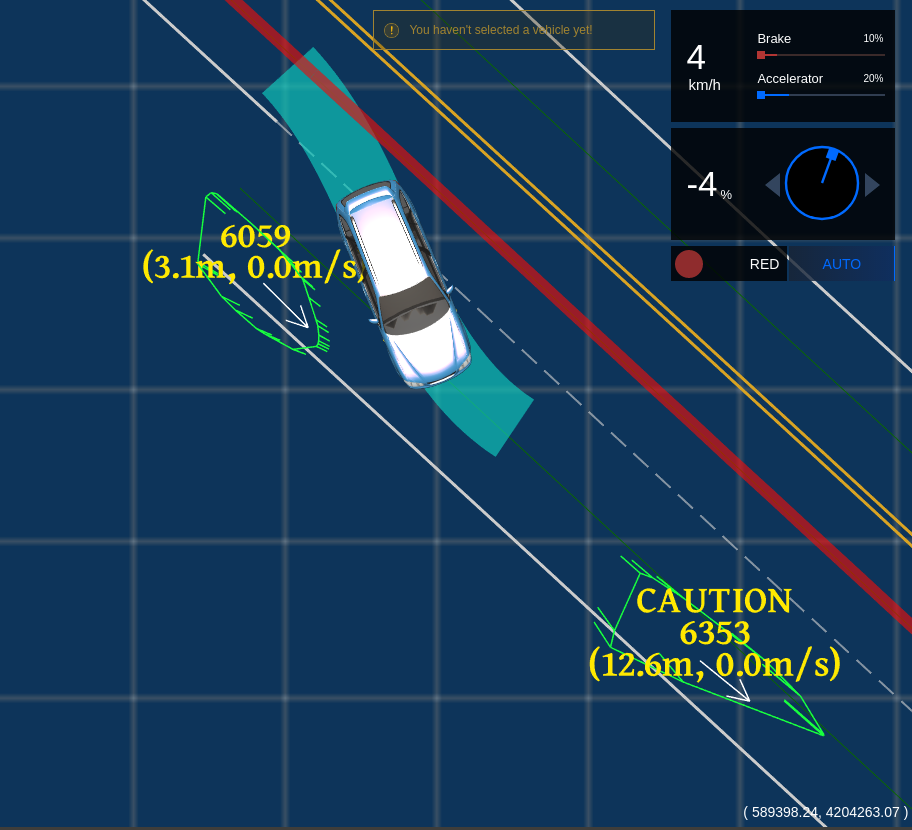}}
\end{minipage}\hfill
 \caption{Pull over scenario at US Field Tests} \label{figure:Pull_over_scenario_at_US_Field_Tests}
\end{figure}

\begin{figure}[ht]
\begin{minipage}{0.16\textwidth}
 \centering
 \subfloat[Beijing T4 Test\label{figure:T4_testing}]{\includegraphics[width=.92\linewidth]{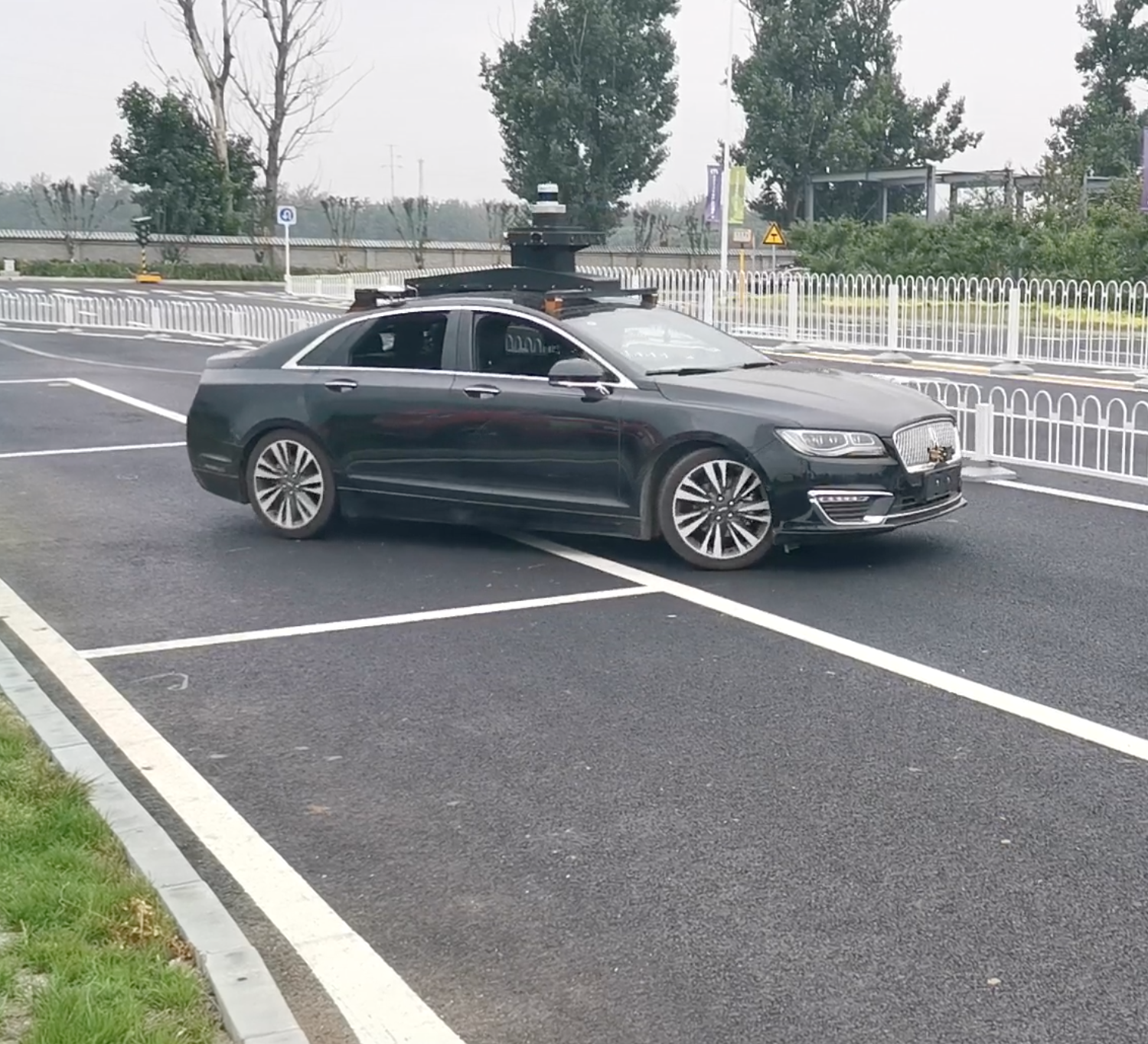}}
 \end{minipage}\hfill
\begin{minipage}{0.16\textwidth}
 \centering
 \subfloat[Test Data: Park In\label{figure:parallel_parking_t4}]{\includegraphics[width=.95\linewidth]{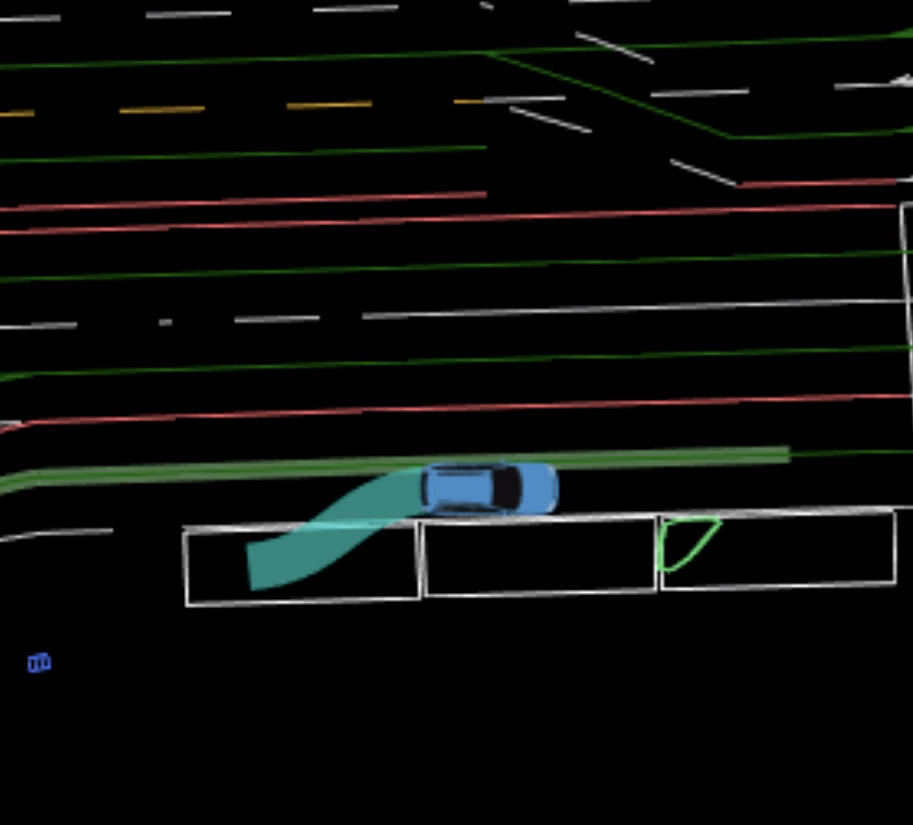}}
 \end{minipage}\hfill
\begin{minipage}{0.16\textwidth}
 \centering
 \subfloat[Test Data: Park Out\label{figure:parallel_parking_stage_parking_out}]{\includegraphics[width=.95\linewidth]{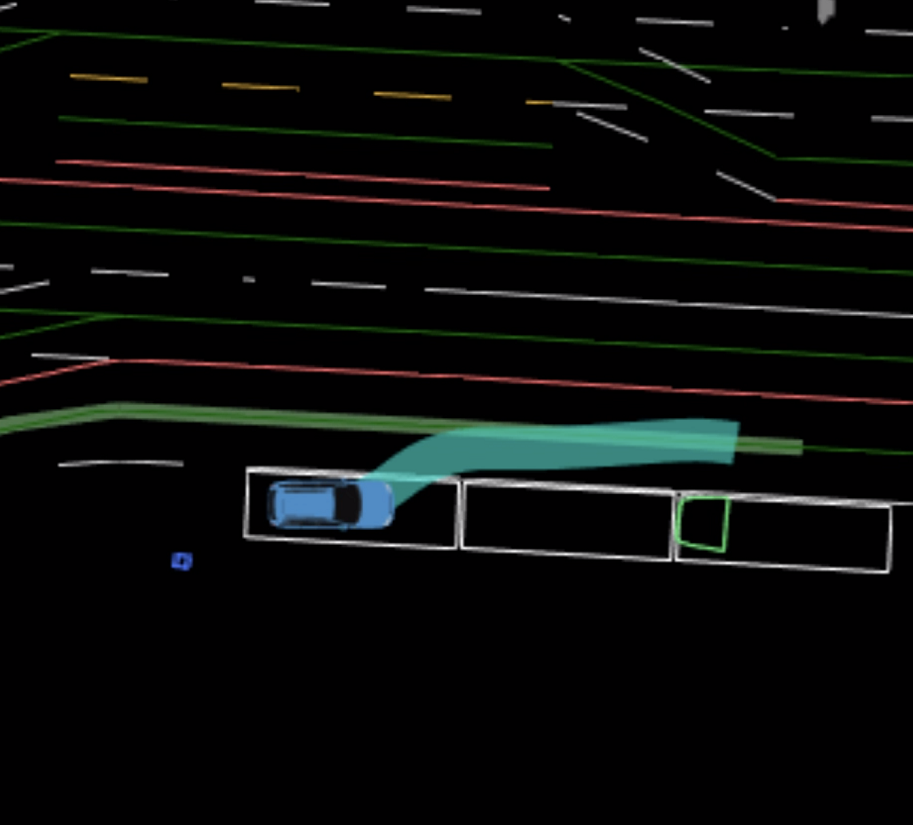}}
 \end{minipage}
 \caption{Parallel parking scenario at China Beijing T4 Tests}
 \label{figure:Parallel_parking_scenario_at_China_Beijing_T4_Tests}
\end{figure}

Our proposed free space planner and controller manipulate the vehicle to handle complex road environment with more subtle maneuvers involving collision avoidance and backward driving. As shown in Fig. \ref{figure:Pull_over_scenario_at_US_Field_Tests} and \ref{figure:Parallel_parking_scenario_at_China_Beijing_T4_Tests}, we have already conducted tests by both US field test and China Beijing T4 test environments. In particular, the China T4 test is currently considered as the most difficult autonomous driving test, due to its strict testing criteria on position/speed precision, robustness and passing rate. This test is derived from the Chinese official guidance document~\cite{t4test} released in 2018 in which the autonomous driving tests are divided into 5 levels from T1 to T5. The higher levels require more complex scenarios and more testing topics. Our planner makes a crucial contribution for us to overcome 10 T4-level free space scenarios, and facilitates Baidu to be the first and so far the only company which passes the entire T4 test in China.

Fig.~\ref{figure:Pull_over_scenario_at_US_Field_Tests} shows the US field test environment, overall optimized planning trajectory, and underway test visualization of the pull over scenario, respectively. 
Fig.~\ref{figure:Parallel_parking_scenario_at_China_Beijing_T4_Tests} show the China Beijing T4 test environment and underway stage-by-stage test visualization of the zig-zag parallel parking scenario, respectively.
Table~\ref{table:t4_test_experimental_data} summarizes the experimental performance data in the pull over and parallel parking tests including multiple forward-driving and backward-driving stages. The high-precision planning and control performance is demonstrated by the very low lateral errors and heading angle errors at every stage through the entire test scenario.

Overall, the experiment results demonstrate that the optimized planning trajectory establishes a kinematic-smooth and kinodynamic-feasible reference for the control module, so as to enable the accurate autonomous vehicle control.

\begin{table}[ht]
\centering
\caption{Experimental performance summary (pull over / parallel parking scenarios with multiple stages)}
\setlength{\tabcolsep}{0.5em}
\begin{tabular}{p{1.4cm} | p{3.0cm} | p{1.5cm} p{1.5cm}}
 \hline
 Test Environment & Test Stages & Lateral Error at End, in (m)  & Heading Error at End, in (deg)\\ 
 \hhline{=:=:=:=}
  & 1. Approaching & 0.0428 & 1.4390\\
 \hhline{~|-|--}
 Beijing T4 & 2. Parking In & 0.0886 & 0.9071\\
 \hhline{~|-|--}
  & 3. Parking Out & 0.0542 & 1.5573\\
 \hhline{~|-|--}
  & \textbf{Mean Value} & \textbf{0.0619} & \textbf{1.3011} \\
 \hhline{=:=:=:=}
  & 1. Approaching & 0.0238 & 1.589\\
 \hhline{~|-|--}
 US Field & 2. Pose Adjustment & 0.0256 & 0.431 \\
 \hhline{~|-|--}
  & 3. Parking In (Backward) & 0.0476 & 4.863\\
 \hhline{~|-|--}
  & 4. Parking In (Forward) & 0.0819 & 0.939\\
 \hhline{~|-|--}
  & \textbf{Mean Value} & \textbf{0.0447} & \textbf{1.956}\\
 \hline
\end{tabular}
\label{table:t4_test_experimental_data}
\end{table}

\section{Conclusion}
 In this paper, we present a novel decoupled trajectory optimization algorithm which has the advantages of real-time computational performance, precise collision avoidance, strict path curvature constraint and comfortable minimum-time speed profile.
Through the exhaustive numeric simulations and real-world autonomous driving test including the US and China Beijing T4 test, we have proved the computation efficiency, control feasibility and robustness of our algorithm. We will extend its applications to other complex scenarios including narrow roads, three point turn, etc.

\bibliographystyle{IEEEtran}
\bibliography{IEEEabrv,./refs}

\end{document}